\pdfoutput=1
\documentclass[11pt]{article}

\usepackage[preprint]{acl}

\usepackage{times}
\usepackage{latexsym}

\usepackage[T1]{fontenc}
\usepackage[utf8]{inputenc}

\usepackage{microtype}

\usepackage{inconsolata}

\usepackage{graphicx}

\usepackage{amssymb}
\usepackage{booktabs}
\usepackage{makecell}
\usepackage{multirow}
\usepackage{xcolor}
\usepackage{xspace}

\newcommand{\CRAC}{CRAC 2026 Shared Task\xspace}
\newcommand{\CRAClong}{CRAC 2026 Shared Task on Multilingual Coreference Resolution\xspace}
\newenvironment{citemize}{\begin{list}{$\bullet$}{\topsep=.1\smallskipamount\itemsep=0pt\parsep=1pt\labelwidth=.5em}}{\end{list}}

\makeatletter\newcommand\notsotiny{\@setfontsize\notsotiny{6.5}{7.5}}\makeatother

\title{CorPipe at CRAC 2026: Empty Nodes and Cross-Lingual Transfer in Multilingual Coreference Resolution}

\author{Milan Straka \\
  Charles University, Faculty of Mathematics and Physics \\
  Institute of Formal and Applied Linguistics \\
  Malostranské nám. 25, Prague, Czech Republic \\
  \texttt{straka@ufal.mff.cuni.cz}
  }

\usepackage{fancyhdr}
\fancypagestyle{officialbibref}{\fancyhf{}\fancyheadoffset[L,R]{50pt}\fancyhead[C]{This paper was published at \textbf{CODI-CRAC 2026} -- please cite the published version {\small\url{https://aclanthology.org/2026.codi-1.27}}.}\fancyfoot[C]{\normalsize\thepage}\addtolength{\topmargin}{-30pt}\addtolength{\headsep}{30pt}}
\begin{document}
\thispagestyle{officialbibref}
\maketitle
\begin{abstract}
  We introduce CorPipe 26, our winning submission to the \CRAClong. The fifth
  edition of this shared task focuses mainly on the comparison of generative
  LLMs and specialized systems; additionally, 5 more datasets and 2 new
  languages are introduced. CorPipe~26 is an improved version of CorPipe~25,
  with a new variant predicting empty nodes together with mentions and
  coreference links in a single model. Our system outperforms all other
  submissions in the LLM track by 2.8 percent points and all submissions in the
  unconstrained track by 9.5 percent points. Furthermore, we perform a series
  of ablation experiments with different model sizes, empty node prediction
  methods, and cross-lingual zero-shot evaluation. The source code and the
  trained models are publicly available at
  {\small\url{https://github.com/ufal/crac2026-corpipe}}.
\end{abstract}

\section{Introduction}

Coreference resolution aims to identify and group together expressions that
refer to the same real-world entity within a text. The
\CRAClong~\citep{novak-etal-2026-findings} is the fifth edition of a shared
task designed to advance multilingual coreference research. Built on the
\hbox{CorefUD 1.4} dataset collection, this year's iteration focuses primarily
on the comparison of generative large language models (LLMs) and specialized systems.
It also expands the task's scope by adding five new datasets and two new
languages (Dutch and Latin), bringing the total to 27 datasets across 19
languages.

As in the previous two years, participating systems must predict \textit{empty
nodes}. These are omitted words that do not appear in the surface text but are
necessary for accurate coreference modeling. Resolving empty nodes is
especially important for pro-drop languages, such as those in the Slavic and
Romance families. In these languages, pronouns are often omitted when they can
be inferred from context, such as through verb morphology. This is shown in the
\hbox{Czech example} \textit{Řekl,~že~nepřijde}, which translates to
\textit{(He) said that (he) won't come}.

Our entry for the \CRAC, CorPipe 26, is an improvement of our past
winning systems~\citep{straka-2025-corpipe,straka-2024-corpipe,straka-2023-ufal,straka-strakova-2022-ufal}.
Our submission is a two-step pipeline: the empty nodes are predicted first
and then mention detection and coreference linking are performed together by
a single model. The empty node prediction system, which has been improved
compared to last year by predicting all available information about the empty
nodes, not just that required for coreference evaluation, has been made
available to all participants as a baseline. All our models are strictly
multilingual (without indicating the language or dataset on the input)
and are trained on all provided datasets simultaneously.

Our main contributions are as follows:
\begin{citemize}
\item We present the top-performing system for the \CRAC, outperforming
  all other submissions in the LLM track by 2.8 percent points and all
  submissions in the unconstrained track by 9.5 percent points.
\item We investigate a one-stage variant of CorPipe 26 predicting empty
  nodes together with mentions and coreference links in a single model,
  surpassing the two-stage variant slightly on the development set.
\item We evaluate the impact of model size, empty node methods, and
  cross-lingual zero-shot settings through ablations.
\item We make the CorPipe 26 source code publicly available under an
  open-source license at {\small\url{https://github.com/ufal/crac2026-corpipe}}.
  Furthermore, we release several pretrained multilingual models of varying
  sizes under the CC BY-NC-SA license.
\end{citemize}

\section{Related Work}

\paragraph{Coreference Resolution}
Most neural coreference models have relied on span-based methods since the work
of \citet{lee-etal-2017-end}, who introduced a model that identifies mentions
and links them simultaneously. \citet{lee-etal-2018-higher} later updated this
approach to make it more efficient and accurate.
\citet{joshi-etal-2020-spanbert} further improved results by using SpanBERT
\citep{joshi-etal-2019-bert}, a model designed specifically to represent text
spans.

Other researchers have explored different designs to avoid the constraints of
span-based methods. \citet{wu-etal-2020-corefqa} treated coreference as
a question-answering task, \citet{liu-etal-2022-autoregressive} developed an
autoregressive system, and \citet{bohnet-etal-2023-coreference} used
a text-to-text approach. A common drawback of these methods, however, is the
computational overhead requiring multiple model calls to process a single sentence.

\paragraph{Word-Level Coreference Resolution}
A major departure from span-based methods came when
\citet{dobrovolskii-2021-word} introduced word-level coreference. Instead of
using whole phrases, this method focuses on the head word of a mention.
\citet{doosterlinck-etal-2023-caw} built on this with CAW-coref to better
handle complex dependencies. More recently, \citet{liu-etal-2024-mscaw}
introduced MSCAW-coref, which supports multiple languages and handles mentions
that appear only once. This word-level approach is now part of Stanza
\citep{qi-etal-2020-stanza}, a popular Python natural language processing
framework.

\paragraph{Multilingual Coreference Resolution}~~
The CRAC shared tasks \citep{zabokrtsky-etal-2022-findings,
zabokrtsky-etal-2023-findings, novak-etal-2024-findings,
novak-etal-2025-findings, novak-etal-2026-findings} have played a central
role in multilingual coreference resolution. They provide a standard evaluation
framework, the CorefUD dataset \citep{CorefUD1.4}, and a multilingual baseline
\citep{prazak-etal-2021-multilingual}.

Earlier versions of our CorPipe system have taken part in every CRAC shared
task. The system has evolved from early multilingual models
\citep{straka-strakova-2022-ufal} to versions that handle wider context
\citep{straka-2023-ufal}, detect zero mentions in raw text
\citep{straka-2024-corpipe}, and a PyTorch version capable of training larger
models~\citep{straka-2025-corpipe}.

\section{Architecture}

CorPipe 26 is based heavily on CorPipe 25. As in the previous version, given
input raw text it starts by predicting the empty nodes using one model and
then performs mention detection and coreference linking together using another
model.

\begin{figure}[t]
    \centering
    \includegraphics[width=.96\hsize]{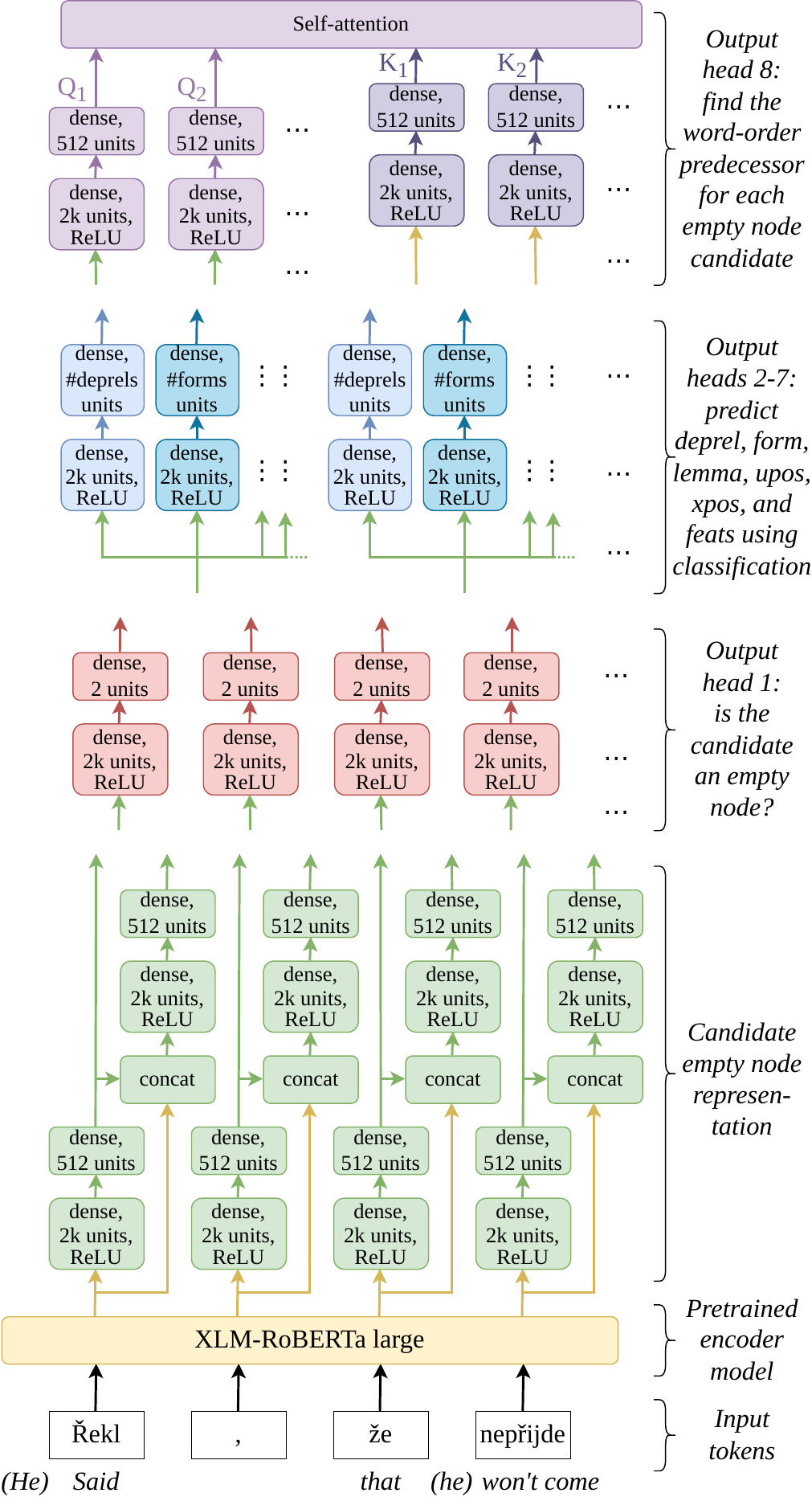}
    \caption{The system architecture of the empty node prediction baseline.
    Every ReLU activation is followed by a dropout layer with a dropout rate of 50\%.}
    \label{fig:empty_nodes_baseline_architecture}
\end{figure}

\begin{table*}[t]
  \footnotesize%
    \setlength{\tabcolsep}{3pt}
  \begin{tabular*}{\hsize}{l*{11}{@{\extracolsep{\fill}}c}}
    \toprule
      Treebank & ARC & DEP & WO & DEP\_WO & FORM & LEMMA & UPOS & XPOS & FEATS & ALL \\
    \midrule
      \texttt{ca}              & 95.55 & 95.55 & 92.74 & 92.74 & ---   & ---   & ---   & ---   & ---   & 92.74 \\
      \texttt{cs\_pcedt}       & 70.91 & 69.36 & 70.77 & 69.21 & 69.07 & 70.77 & 70.91 & 70.91 & 68.08 & 67.80 \\
      \texttt{cs\_pdt}         & 79.21 & 78.34 & 78.52 & 78.00 & 77.83 & 78.00 & 79.03 & 79.21 & 77.14 & 76.79 \\
      \texttt{cs\_pdtsc}       & 85.88 & 85.11 & 85.16 & 84.39 & 84.59 & 85.73 & 85.88 & 85.88 & 82.79 & 81.91 \\
      \texttt{cu}              & 80.55 & 79.50 & 79.77 & 78.85 & ---   & ---   & ---   & ---   & ---   & 78.85 \\
      \texttt{es}              & 95.74 & 95.74 & 93.48 & 93.48 & ---   & ---   & ---   & ---   & ---   & 93.48 \\
      \texttt{grc}             & 89.85 & 87.90 & 89.85 & 87.90 & ---   & ---   & ---   & ---   & ---   & 87.90 \\
      \texttt{hu\_korkor}      & 85.44 & 79.61 & 83.50 & 77.67 & 85.44 & ---   & ---   & ---   & ---   & 77.67 \\
      \texttt{hu\_szegedkoref} & 92.48 & 89.86 & 91.82 & 89.20 & ---   & ---   & ---   & ---   & ---   & 89.20 \\
      \texttt{pl}              & 90.99 & 90.88 & 90.88 & 90.76 & ---   & ---   & ---   & ---   & ---   & 90.65 \\
      \texttt{tr}              & 84.82 & 84.82 & 84.72 & 84.72 & 82.26 & ---   & ---   & ---   & ---   & 82.17 \\
    \bottomrule
  \end{tabular*}

  \caption{Empty nodes prediction baseline performance on the minidev sets of the CRAC 2026 dataset containing empty nodes. Each reported
    metric is an F1 score where a prediction is considered correct if both the
    dependency head and the given column are correct, with ARC denoting the dependency head, DEP the dependency relation, WO the word order position,
    and ALL the combination of all other predictions.}
  \label{tab:empty_nodes}
\end{table*}

\paragraph{Empty Nodes Baseline}
The empty nodes are predicted using our baseline system that was provided
to all shared task participants. It is a substantial improvement over the
baseline provided in previous years, which produced only the information
necessary for coreference resolution evaluation (word order position,
dependency parent, and dependency relation). We have extended the system to
predict all available empty node information, additionally providing the form,
lemma, UPOS, XPOS, and FEATS columns. The model operates non-autoregressively,
predicting up to two empty nodes per input word, with each input word serving
as a potential dependency head.

Figure~\ref{fig:empty_nodes_baseline_architecture} presents the architecture of
this system. The input words of a single sentence are tokenized and
processed through an XLM-RoBERTa-large encoder~\citep{conneau-etal-2020-unsupervised},
with each input word represented by its first subword embedding. For each word, we generate two empty node
candidates: the first through a dense-ReLU-dropout-dense module
(768$\rightarrow2k\rightarrow$512 units), and the second by concatenating the
first candidate with the input word representation and applying an analogous
transformation. The candidates are processed by eight heads, each first passing
its input through its own 2k-unit ReLU layer and dropout: (1) a binary
classification head determining whether the candidate will produce an empty
node, (2-7) classification heads predicting the dependency relation, form,
lemma, UPOS, XPOS, and FEATS, and (8) a self-attention word-order
prediction head identifying the insertion point for the empty node. Please
refer to the source code for more details.

We train a single multilingual model on a concatenation of all
corpora containing empty nodes, with sentences sampled proportionally to the
square root of their respective corpus sizes. We use the Adam
optimizer~\citep{kingma-and-ba-2015-adam} and train for 20 epochs of 5\,000 batches,
each consisting of 64 sentences. The learning rate linearly increases to 1e-5 in the first
epoch and then decays to zero following cosine decay~\citep{loshchilov-etal-2017-sgdr}.
The intrinsic performance of the system is summarized in
Table~\ref{tab:empty_nodes}.

The source code of the system is available at {\small\url{https://github.com/ufal/crac2026_empty_nodes_baseline}}
under the open-source MPL license. The trained model is available both on
Hugging Face and via LINDAT/CLARIAH-CZ. Finally, the minidev and minitest sets
of the \CRAC with predicted empty nodes are available to all participants.

As a part of the shared task submission, we also trained an improved version of
the empty node prediction system by using a larger batch size of 384 sentences
(the maximum batch size allowed by a single H100 GPU), compared to the 96 sentences used by the baseline.
The intrinsic evaluation of this improved system is presented in
Table~\ref{tab:empty_nodes_improved} in the Appendix. Overall, the improved
system provides a boost of 2.2 percent points in the DEP metric and 2.5 percent
points in the ALL metric.

\begin{figure*}[t]
    \centering
    \includegraphics[width=.9\hsize]{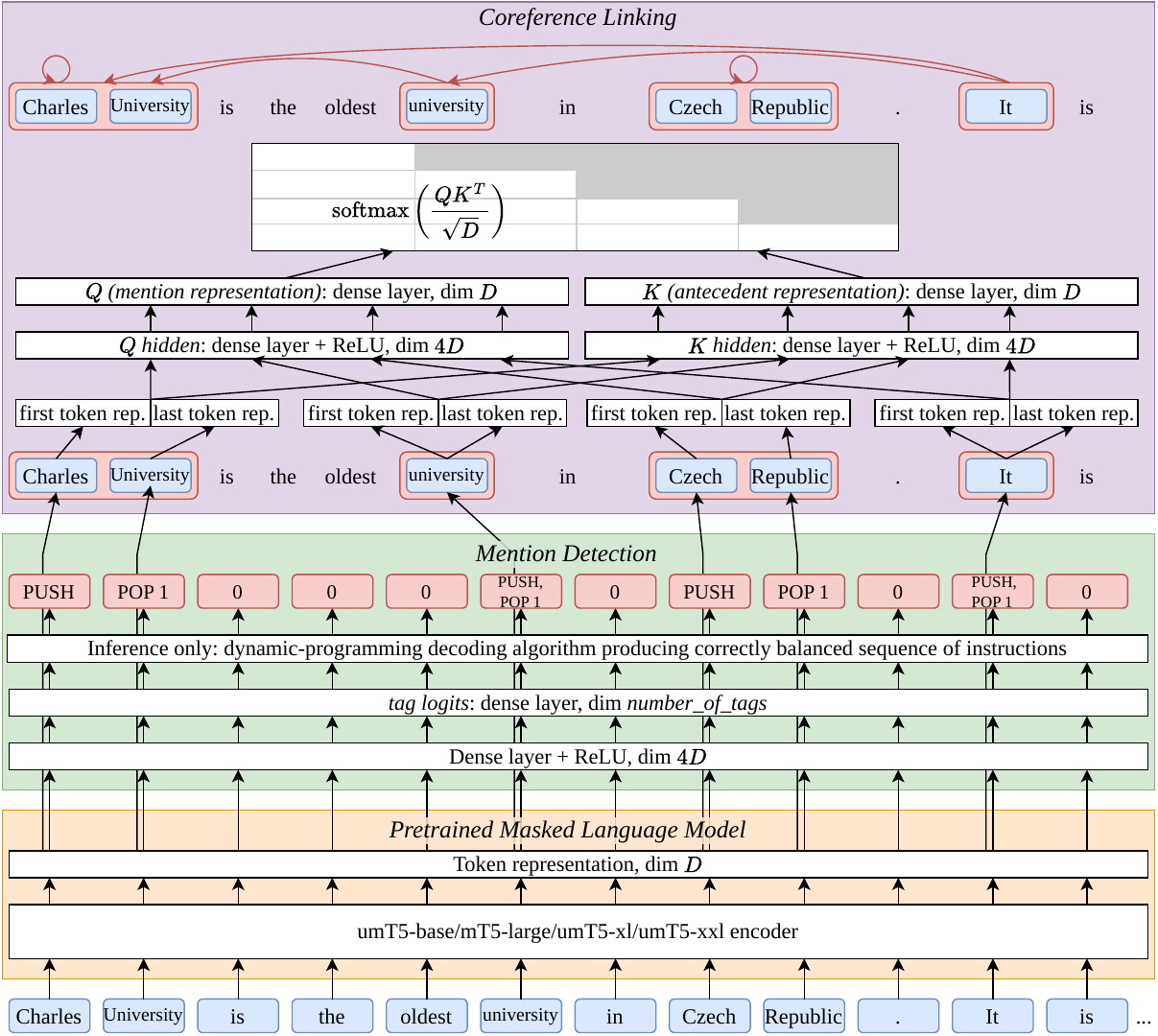}
    \caption{The CorPipe 26 model architecture.}
    \label{fig:corpipe26_architecture}
\end{figure*}

\paragraph{Coreference Resolution}
With the empty nodes predicted, we perform coreference resolution by predicting
mentions and coreference links together using a single model. The architecture
of this model is the same as the one used in CorPipe 25~\citep{straka-2025-corpipe}
and it is presented in Figure~\ref{fig:corpipe26_architecture} and summarized
below; see the referenced work for a detailed description.

The model processes documents one sentence at a time. To include as much
context as possible, we expand every input sentence with preceding tokens and
at most 50 following tokens to the extent allowed by the maximum segment length
(512 or 2\,560 tokens). The input tokens are passed through a pretrained
multilingual encoder, and coreference mentions are predicted using an enhanced
BIO encoding scheme capable of representing potentially overlapping spans.
The identified mentions are then represented as a concatenation of their
first and last tokens, and coreference links are predicted through
a self-attention layer determining the most likely antecedent for each mention,
allowing self-reference to indicate the first mention of an entity.

\looseness-1
We employ different segment sizes during training (512 tokens) and
inference (2\,560 tokens, except for two PROIEL corpora using 512
tokens) to improve modeling of long-range coreference links; this difference
is allowed by the relative positional encodings used by our pretrained
encoders.

\begin{table}[t]
  \setlength{\tabcolsep}{4.5pt}
  \begin{tabular*}{\hsize}{lrccc}
    \toprule
      Model & Params & \makecell[c]{Batch\\Size} & \makecell[c]{Learning\\Rate} & \makecell{Train\\Time} \\
    \midrule
      umT5 base &  269M & 8 & 6e-4 & \hphantom{0}5h \\
      mT5 large &  538M & 8 & 6e-4 & 10h\\
      umT5 xl   & 1605M & 6 & 5e-4 & 22h\\
      umT5 xxl  & 5417M & 6 & 5e-4 & 36h \\
    \bottomrule
  \end{tabular*}
  \caption{Properties of encoder models used. The training time is
  measured for 15 epochs 10k updates each using a single A100 GPU, with the
  exception of the xxl models, which are trained using a single H100 GPU.}
  \label{tab:mt5_model_variants}
\end{table}

We consider four sizes of the pretrained encoders:
umT5-base~\citep{chung-etal-2023-unimax},
mT5-large~\citep{xue-etal-2021-mt5}, umT5-xl, and umT5-xxl.
For each encoder, we train 10 models differing only in random initialization.
Each model is trained for 15 epochs each consisting of 10k batches using
the AdaFactor optimizer~\citep{shazeer-etal-2018-adafactor}. The learning rate
first linearly increases during the initial 10\% of training and then decays
to zero following cosine decay~\citep{loshchilov-etal-2017-sgdr}. The
parameter counts of the models, as well as the learning rate, batch size, and
training time, are summarized in Table~\ref{tab:mt5_model_variants}.

\begin{figure}[t]
    \centering
    \includegraphics[width=\hsize]{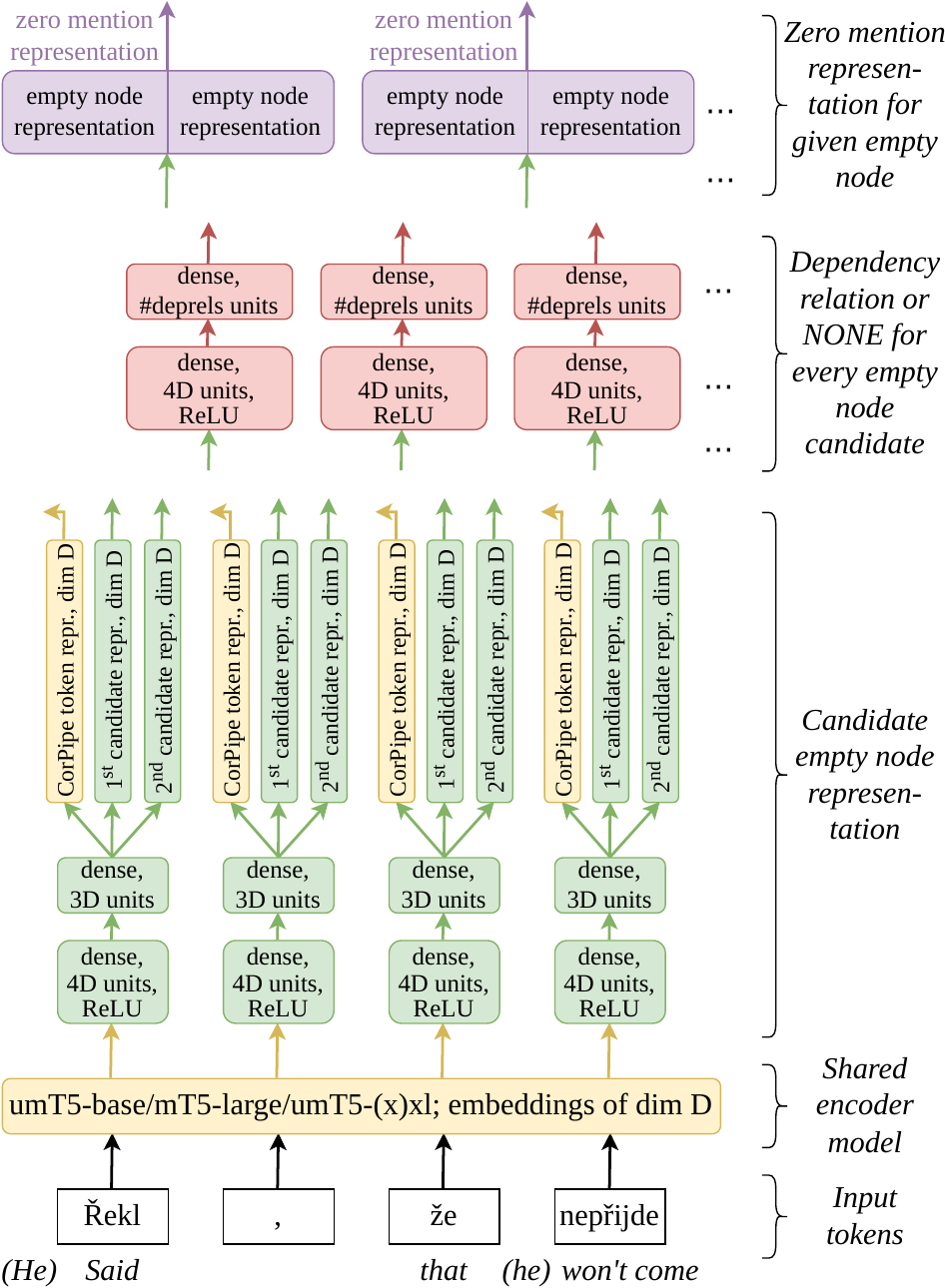}
    \caption{The modifications of the one-stage CorPipe 26 variant to the standard
    CorPipe 26 architecture.}
    \label{fig:corpipe26_1stage_architecture}
\end{figure}

\paragraph{One-Stage Variant}
In addition to the described CorPipe 26, we also evaluate a one-stage variant
predicting the empty nodes together with mentions and coreference links in
a single model. The architecture is inspired by the single-stage
CorPipe~24~\citep{straka-2024-corpipe}, with a few modifications improving
the performance considerably.

Unlike the empty node baseline, the one-stage variant does not predict every
empty node or all of its details. It only predicts empty nodes that are
actually coreference mentions, providing for each just the dependency head and
relation, the minimum needed for coreference evaluation. Additionally, each
predicted empty node is treated as a full mention. While this simplification
may reduce the exact-match score, it does not affect the head-match score
metric.

The additions of the one-stage variant to the standard CorPipe 26 architecture
are presented in Figure~\ref{fig:corpipe26_1stage_architecture}. The encoder
embeddings of size $D$ are first passed through a $4D$-unit ReLU layer and
dropout followed by a $3D$ dense layer that is split into three $D$-dimensional
embeddings for (a) the token representation used by the mention detection and
coreference linking heads, and (b) two empty node candidates. All empty node
candidates are processed by a shared $4D$-unit ReLU layer, dropout, and a
classification layer predicting either \verb|NONE|, indicating the candidate
is not an empty node, or the dependency relation of the empty node, whose
head is then the input token. The zero mentions are constructed by repeating
the candidate embeddings twice; they are added to the mentions predicted
from the input tokens, and coreference linking is performed as in the standard
CorPipe 26.

We train the variant using the same encoders and hyperparameters as the
standard CorPipe 26.

\begin{table}[t]
    \footnotesize\setlength{\tabcolsep}{3pt}
  \renewcommand\cellset{\renewcommand\arraystretch{0.85}}
  \catcode`@ = 13\def@{\bfseries}
  \begin{tabular*}{\hsize}{l*{4}{@{\extracolsep{\fill}}c}}
    \toprule
      System & \makecell{Head-\\match} & \makecell{Partial-\\match} & \makecell{Exact-\\match} & \makecell{With Sin-\\gletons} \\
  \midrule
  \multicolumn{5}{l}{\textsc{Unconstrained}} \\[3pt]
    ~~@CorPipeEnsemble       &@\makecell[c]{77.11 \\ 1} &@\makecell[c]{76.30 \\ 1} &@\makecell[c]{74.07 \\ 1} &@\makecell[c]{79.11 \\ 1} \\[7pt]
    ~~@CorPipeXXL            & \makecell[c]{76.18 \\ 2} & \makecell[c]{75.32 \\ 2} & \makecell[c]{72.90 \\ 2} & \makecell[c]{78.15 \\ 2} \\[7pt]
    ~~@CorPipeLarge          & \makecell[c]{72.32 \\ 3} & \makecell[c]{71.16 \\ 3} & \makecell[c]{68.79 \\ 3} & \makecell[c]{74.52 \\ 3} \\[7pt]
    ~~DAggerCoref            & \makecell[c]{67.56 \\ 4} & \makecell[c]{67.56 \\ 4} & \makecell[c]{37.63 \\ 7} & \makecell[c]{58.75 \\ 5} \\[7pt]
    ~~Stanza                 & \makecell[c]{67.00 \\ 5} & \makecell[c]{65.87 \\ 5} & \makecell[c]{63.32 \\ 4} & \makecell[c]{68.53 \\ 4} \\[7pt]
    ~~\textsc{Baseline}-GZ   & \makecell[c]{55.39 \\ 6} & \makecell[c]{55.06 \\ 6} & \makecell[c]{53.91 \\ 5} & \makecell[c]{48.28 \\ 6} \\[7pt]
    ~~\textsc{Baseline}      & \makecell[c]{54.54 \\ 7} & \makecell[c]{54.16 \\ 7} & \makecell[c]{52.96 \\ 6} & \makecell[c]{47.45 \\ 7} \\[7pt]
    ~~AUKBC-MULCRF           & \makecell[c]{35.24 \\ 8} & \makecell[c]{35.21 \\ 8} & \makecell[c]{20.65 \\ 8} & \makecell[c]{33.10 \\ 8} \\

  \midrule
  \multicolumn{5}{l}{\textsc{LLM}} \\[3pt]
    ~~LLM-LatticeNLP         &@\makecell[c]{74.32 \\ 1} &@\makecell[c]{74.32 \\ 1} & \makecell[c]{41.83 \\ 2} &@\makecell[c]{76.09 \\ 1} \\[7pt]
    ~~LLM-UWB                & \makecell[c]{73.83 \\ 2} & \makecell[c]{73.83 \\ 2} & \makecell[c]{40.76 \\ 3} & \makecell[c]{75.59 \\ 2} \\[7pt]
    ~~LLM-PortNLP            & \makecell[c]{68.69 \\ 3} & \makecell[c]{67.51 \\ 3} &@\makecell[c]{65.30 \\ 1} & \makecell[c]{70.98 \\ 3} \\[7pt]
    ~~LLM-Landcore           & \makecell[c]{46.19 \\ 4} & \makecell[c]{44.79 \\ 4} & \makecell[c]{40.47 \\ 4} & \makecell[c]{47.78 \\ 4} \\

  \bottomrule
\end{tabular*}

  \caption{Official results of \CRAC~on the minitest set with various metrics in \%.}
  \label{tab:official_metrics}
\end{table}

\section{Shared Task Results}

We submitted the maximum number of three systems to the \CRAC, all based on the
standard two-stage CorPipe 26 architecture (unfortunately, we did not finish training
the one-stage models before the submission deadline):
\begin{citemize}
  \item \textbf{CorPipeLarge}, a single best-performing model according to the
    minidev set performance based on the mT5-large encoder;
  \item \textbf{CorPipeXXL}, using the best model with the umT5-xxl encoder and
    improved empty node prediction system;
  \item \textbf{CorPipeEnsemble}, an ensemble of the best 7 out of 10 umT5-xxl models, again
    utilizing the improved empty node prediction system.
\end{citemize}

\begin{table*}[t!]
    \notsotiny\setlength{\tabcolsep}{3pt}
  \renewcommand\cellset{\renewcommand\arraystretch{0.70}}
  \catcode`@ = 13\def@{\bfseries}
\begin{tabular*}{\hsize}{l*{28}{@{\extracolsep{\fill}}c}}
    \toprule
      System & Avg &
      \texttt{ca} &
      \makecell[c]{\texttt{cs} \\ \texttt{\kern-.1em pced\kern-.1em}} &
      \makecell[c]{\texttt{cs} \\ \texttt{pdt}} &
      \makecell[c]{\texttt{cs} \\ \texttt{\kern-.1em pdts\kern-.1em}} &
      \texttt{cu} &
      \texttt{de} &
      \makecell[c]{\texttt{en} \\ \texttt{\kern-.1em fant\kern-.1em}} &
      \makecell[c]{\texttt{en} \\ \texttt{gum}} &
      \makecell[c]{\texttt{en} \\ \texttt{\kern-.1em litb\kern-.1em}} &
      \texttt{es} &
      \makecell[c]{\texttt{fr} \\ \texttt{\kern-.1em anco\kern-.1em}} &
      \makecell[c]{\texttt{fr} \\ \texttt{\kern-.1em demo\kern-.1em}} &
      \makecell[c]{\texttt{fr} \\ \texttt{\kern-.1em litb\kern-.1em}} &
      \texttt{grc} &
      \texttt{hbo} &
      \texttt{hi} &
      \makecell[c]{\texttt{hu} \\ \texttt{\kern-.1em kork\kern-.1em}} &
      \makecell[c]{\texttt{hu} \\ \texttt{\kern-.1em szeg\kern-.1em}} &
      \texttt{ko} &
      \texttt{la} &
      \texttt{lt} &
      \texttt{nl} &
      \makecell[c]{\texttt{no} \\ \texttt{\kern-.1em bokm\kern-.1em}} &
      \makecell[c]{\texttt{no} \\ \texttt{\kern-.1em nyno\kern-.1em}} &
      \texttt{pl} &
      \texttt{ru} &
      \texttt{tr} \\
  \midrule
  \multicolumn{5}{l}{\textsc{Unconstrained}} \\
\bottomrule
\end{tabular*}

  \caption{Official results of \CRAC~on the minitest set (CoNLL score in \%). The systems $^\dagger$ are described in \citet{prazak-etal-2021-multilingual}; the rest in \citet{novak-etal-2026-findings}.}
  \label{tab:official_treebanks}
\end{table*}

\begin{table*}[t!]
  
    \notsotiny\setlength{\tabcolsep}{1pt}
    \catcode`@ = 13\def@{\bfseries}
    \catcode`! = 13\def!{\itshape}
    \begin{tabular*}{\hsize}{l*{28}{@{\extracolsep{\fill}}c}}
      \toprule
        System & Avg &
        \texttt{ca} &
        \makecell[c]{\texttt{cs} \\ \texttt{\kern-.1em pced\kern-.1em}} &
        \makecell[c]{\texttt{cs} \\ \texttt{pdt}} &
        \makecell[c]{\texttt{cs} \\ \texttt{\kern-.1em pdts\kern-.1em}} &
        \texttt{cu} &
        \texttt{de} &
        \makecell[c]{\texttt{en} \\ \texttt{\kern-.1em fant\kern-.1em}} &
        \makecell[c]{\texttt{en} \\ \texttt{gum}} &
        \makecell[c]{\texttt{en} \\ \texttt{\kern-.1em litb\kern-.1em}} &
        \texttt{es} &
        \makecell[c]{\texttt{fr} \\ \texttt{\kern-.1em anco\kern-.1em}} &
        \makecell[c]{\texttt{fr} \\ \texttt{\kern-.1em demo\kern-.1em}} &
        \makecell[c]{\texttt{fr} \\ \texttt{\kern-.1em litb\kern-.1em}} &
        \texttt{grc} &
        \texttt{hbo} &
        \texttt{hi} &
        \makecell[c]{\texttt{hu} \\ \texttt{\kern-.1em kork\kern-.1em}} &
        \makecell[c]{\texttt{hu} \\ \texttt{\kern-.1em szeg\kern-.1em}} &
        \texttt{ko} &
        \texttt{la} &
        \texttt{lt} &
        \texttt{nl} &
        \makecell[c]{\texttt{no} \\ \texttt{\kern-.1em bokm\kern-.1em}} &
        \makecell[c]{\texttt{no} \\ \texttt{\kern-.1em nyno\kern-.1em}} &
        \texttt{pl} &
        \texttt{ru} &
        \texttt{tr} \\
    \midrule
\multicolumn{22}{l}{\textsc{A) CorPipe Single Models}} \\
\bottomrule
    \end{tabular*}

  \caption{Additional experiments on the CorefUD 1.4 minitest set (CoNLL score in \%).}
  \label{tab:test_ablations}
\end{table*}

\begin{table*}[t!]
  
    \notsotiny\setlength{\tabcolsep}{1pt}
    \catcode`@ = 13\def@{\bfseries}
    \catcode`! = 13\def!{\itshape}
    \begin{tabular*}{\hsize}{l*{28}{@{\extracolsep{\fill}}c}}
      \toprule
        System & Avg &
        \texttt{ca} &
        \makecell[c]{\texttt{cs} \\ \texttt{\kern-.1em pced\kern-.1em}} &
        \makecell[c]{\texttt{cs} \\ \texttt{pdt}} &
        \makecell[c]{\texttt{cs} \\ \texttt{\kern-.1em pdts\kern-.1em}} &
        \texttt{cu} &
        \texttt{de} &
        \makecell[c]{\texttt{en} \\ \texttt{\kern-.1em fant\kern-.1em}} &
        \makecell[c]{\texttt{en} \\ \texttt{gum}} &
        \makecell[c]{\texttt{en} \\ \texttt{\kern-.1em litb\kern-.1em}} &
        \texttt{es} &
        \makecell[c]{\texttt{fr} \\ \texttt{\kern-.1em anco\kern-.1em}} &
        \makecell[c]{\texttt{fr} \\ \texttt{\kern-.1em demo\kern-.1em}} &
        \makecell[c]{\texttt{fr} \\ \texttt{\kern-.1em litb\kern-.1em}} &
        \texttt{grc} &
        \texttt{hbo} &
        \texttt{hi} &
        \makecell[c]{\texttt{hu} \\ \texttt{\kern-.1em kork\kern-.1em}} &
        \makecell[c]{\texttt{hu} \\ \texttt{\kern-.1em szeg\kern-.1em}} &
        \texttt{ko} &
        \texttt{la} &
        \texttt{lt} &
        \texttt{nl} &
        \makecell[c]{\texttt{no} \\ \texttt{\kern-.1em bokm\kern-.1em}} &
        \makecell[c]{\texttt{no} \\ \texttt{\kern-.1em nyno\kern-.1em}} &
        \texttt{pl} &
        \texttt{ru} &
        \texttt{tr} \\
    \midrule
\multicolumn{22}{l}{\textsc{A) Submitted CorPipe Single Models}} \\
\bottomrule
    \end{tabular*}

  \caption{Ablation experiments on the CorefUD 1.4 minidev set (CoNLL score in \%). The results of submitted models are averages of 7 runs, other results are averages of 2 or more runs.}
  \label{tab:dev_ablations}
\end{table*}

\begin{table*}[t!]
  \renewcommand\arraystretch{1.08}
  
    \notsotiny\setlength{\tabcolsep}{1pt}
    \catcode`@ = 13\def@{\bfseries}
    \catcode`! = 13\def!{\itshape}
    \begin{tabular*}{\hsize}{l*{28}{@{\extracolsep{\fill}}c}}
      \toprule
        System & Avg &
        \texttt{ca} &
        \makecell[c]{\texttt{cs} \\ \texttt{\kern-.1em pced\kern-.1em}} &
        \makecell[c]{\texttt{cs} \\ \texttt{pdt}} &
        \makecell[c]{\texttt{cs} \\ \texttt{\kern-.1em pdts\kern-.1em}} &
        \texttt{cu} &
        \texttt{de} &
        \makecell[c]{\texttt{en} \\ \texttt{\kern-.1em fant\kern-.1em}} &
        \makecell[c]{\texttt{en} \\ \texttt{gum}} &
        \makecell[c]{\texttt{en} \\ \texttt{\kern-.1em litb\kern-.1em}} &
        \texttt{es} &
        \makecell[c]{\texttt{fr} \\ \texttt{\kern-.1em anco\kern-.1em}} &
        \makecell[c]{\texttt{fr} \\ \texttt{\kern-.1em demo\kern-.1em}} &
        \makecell[c]{\texttt{fr} \\ \texttt{\kern-.1em litb\kern-.1em}} &
        \texttt{grc} &
        \texttt{hbo} &
        \texttt{hi} &
        \makecell[c]{\texttt{hu} \\ \texttt{\kern-.1em kork\kern-.1em}} &
        \makecell[c]{\texttt{hu} \\ \texttt{\kern-.1em szeg\kern-.1em}} &
        \texttt{ko} &
        \texttt{la} &
        \texttt{lt} &
        \texttt{nl} &
        \makecell[c]{\texttt{no} \\ \texttt{\kern-.1em bokm\kern-.1em}} &
        \makecell[c]{\texttt{no} \\ \texttt{\kern-.1em nyno\kern-.1em}} &
        \texttt{pl} &
        \texttt{ru} &
        \texttt{tr} \\
    \midrule
\multicolumn{22}{l}{\textsc{A) CorPipe One-stage Single Models}} \\
\bottomrule
    \end{tabular*}

  \caption{Two- and one-stage post-competition comparison on the CorefUD 1.4 minitest set (CoNLL score in \%).}
  \label{tab:test_twostage_onestage_ablations}
\end{table*}

The official results of the \CRAC are presented in
Table~\ref{tab:official_metrics} showing four minitest metrics of all the submitted
systems, and in Table~\ref{tab:official_treebanks} displaying the minitest
CoNLL scores of all individual treebanks.
CorPipeXXL and CorPipeEnsemble outperform all other submissions, both
in the LLM track and the unconstrained track, by large margins of 2.8 and 9.5
percent points for the best CorPipeEnsemble model, respectively.

\looseness-1
Table~\ref{tab:test_ablations} presents the performance of all four model sizes
as well as model ensembles of all sizes on the minitest set. The model
performance increases reliably with the model size, both for single
models and ensembles, resulting in increases of circa +3, +2.5, and +3.5
percent points for gradually increasing model sizes. Furthermore, the ensembles
improve the performance by circa 0.9 percent points on average. Lastly, the
improved empty node prediction system provides a modest boost of 0.23 percent
points for the xxl-sized models.

\paragraph{Ablation Experiments}
Table~\ref{tab:dev_ablations} presents our ablation experiments on the
minidev sets, starting with the mean performance of the 7 CorPipe~26
models of all sizes used for official ensemble submission shown in Section A.
The one-stage variants for all four model sizes are evaluated in Section B
and compared to the standard two-stage variants in Section C. The one-stage
variants outperform the two-stage variants for all model sizes (even if
the best 7 out of 10 two-stage models are reported), by an average
of 0.25 percent points, demonstrating the effectiveness of the proposed
one-stage variant architecture. Note that in CorPipe 24, the one-stage variant
was outperformed by the two-stage variant by more than 1 percent point.

While we did not finish training the one-stage models before the shared task
deadline, the organizers enabled us to perform a post-competition evaluation of our
one-stage models on the minitest sets. The resulting comparison with the two-stage models
is presented in Table~\ref{tab:test_twostage_onestage_ablations}. While
the relative improvements of the one-stage models are less consistent on the
minitest sets, the xxl-sized one-stage model outperforms the two-stage model
both with the original and the improved empty node prediction system. Finally,
note that the evaluated one-stage models are the best out of 2 models trained,
compared to the best out of 10 models for the two-stage variant.

\paragraph{Zero-Shot Cross-Lingual Transfer}
\looseness1
Section D of Table~\ref{tab:dev_ablations} quantifies the zero-shot
cross-lingual transfer of the models. When showcasing the performance on
a dataset in a given language, we train a model on all datasets except those in
the same language. Considering the one-stage variants, the average drop in
performance across all datasets and model sizes is 17.9 percent points. While
the performance drop is large, both the large-sized and the xl-sized one-stage
zero-shot models still outperform the baseline system provided by the
\hbox{organizers}.

To evaluate the zero-shot performance of the two-stage model, we train both the
empty node baseline and the coreference resolution model in a zero-shot manner.
The average drop in this zero-shot two-stage setting is marginally larger, 18.6
percent points on average. For comparison, we also evaluate the zero-shot
performance of the two-stage model when using the provided empty node
prediction baseline, which is applied only on the datasets containing empty
nodes. The resulting performance is 3.6 percent points higher than the
zero-shot two-stage model, still significantly worse than the non-zero-shot
model.

Comparing the zero-shot performances across languages, when a closely related
language (maybe with a similar annotation scheme) is still present in the
training data, like for Catalan when Spanish is still present and vice versa,
the performance drop is much smaller, around 3-5 percent points. Furthermore,
we hypothesize that the performance drop depends considerably on the
annotation scheme, as suggested by the zero-shot performance on French, where
the three datasets demonstrate performance decrease of -10, -20, and -30
percent points, respectively, indicating that it is not just the language that
affects the zero-shot performance.

\begin{table*}[t!]
  
    \notsotiny\setlength{\tabcolsep}{5pt}
    \catcode`@ = 13\def@{\bfseries}
    \catcode`! = 13\def!{\itshape}
    \begin{tabular*}{\hsize}{l*{13}{@{\extracolsep{\fill}}c}}
      \toprule
        System & Avg &
        \texttt{ca} &
        \makecell[c]{\texttt{cs} \\ \texttt{\kern-.1em pced\kern-.1em}} &
        \makecell[c]{\texttt{cs} \\ \texttt{pdt}} &
        \makecell[c]{\texttt{cs} \\ \texttt{\kern-.1em pdts\kern-.1em}} &
        \texttt{cu} &
        \texttt{es} &
        \texttt{grc} &
        \makecell[c]{\texttt{hu} \\ \texttt{\kern-.1em kork\kern-.1em}} &
        \makecell[c]{\texttt{hu} \\ \texttt{\kern-.1em szeg\kern-.1em}} &
        \texttt{pl} &
        \texttt{tr} \\
    \midrule
\multicolumn{13}{l}{\textsc{A) Empty Node Prediction Ablations for mT5-large Model}} \\[2pt]
~~Empty node baseline, single mT5-large model & \textcolor{black}{71.59} & \textcolor{black}{81.3} & \textcolor{black}{74.4} & \textcolor{black}{76.9} & \textcolor{black}{72.0} & \textcolor{black}{58.3} & \textcolor{black}{82.0} & \textcolor{black}{69.6} & \textcolor{black}{65.8} & \textcolor{black}{68.7} & \textcolor{black}{76.4} & \textcolor{black}{62.2} \\[.5pt]
~~\multirow{2}{*}{No empty nodes, single mT5-large model} & \textcolor{red!100.0!black}{--\kern 0.04em 11.8} & \textcolor{red!100.0!black}{--\kern 0.04em 6.1} & \textcolor{red!100.0!black}{--\kern 0.04em 8.3} & \textcolor{red!100.0!black}{--\kern 0.04em 4.8} & \textcolor{red!100.0!black}{--\kern 0.04em 16} & \textcolor{red!100.0!black}{--\kern 0.04em 12} & \textcolor{red!100.0!black}{--\kern 0.04em 5.8} & \textcolor{red!100.0!black}{--\kern 0.04em 16} & \textcolor{red!100.0!black}{--\kern 0.04em 13} & \textcolor{red!100.0!black}{--\kern 0.04em 6.7} & \textcolor{red!100.0!black}{--\kern 0.04em 11} & \textcolor{red!100.0!black}{--\kern 0.04em 30} \\[-1pt]
~~ & \textcolor{black}{59.84} & \textcolor{black}{75.2} & \textcolor{black}{66.1} & \textcolor{black}{72.1} & \textcolor{black}{55.9} & \textcolor{black}{46.2} & \textcolor{black}{76.2} & \textcolor{black}{54.4} & \textcolor{black}{53.2} & \textcolor{black}{62.0} & \textcolor{black}{64.7} & \textcolor{black}{32.3} \\[.5pt]
~~\multirow{2}{*}{One-stage mT5-large model} & \textcolor{blue!18.9!black}{+0.67} & \textcolor{blue!68.4!black}{+0.6} & \textcolor{blue!29.7!black}{+1.3} & \textcolor{blue!65.4!black}{+1.0} & \textcolor{blue!13.6!black}{+0.6} & \textcolor{blue!17.1!black}{+1.0} & \textcolor{blue!81.1!black}{+0.6} & \textcolor{blue!58.8!black}{+2.9} & \textcolor{red!1.0!black}{--\kern 0.04em 0.1} & \textcolor{red!9.2!black}{--\kern 0.04em 0.6} & \textcolor{red!0.1!black}{+0.0} & \textcolor{blue!1.5!black}{+0.2} \\[-1pt]
~~ & \textcolor{black}{72.26} & \textcolor{black}{81.9} & \textcolor{black}{75.7} & \textcolor{black}{77.9} & \textcolor{black}{72.5} & \textcolor{black}{59.3} & \textcolor{black}{82.6} & \textcolor{black}{72.5} & \textcolor{black}{65.7} & \textcolor{black}{68.1} & \textcolor{black}{76.4} & \textcolor{black}{62.4} \\[.5pt]
~~\multirow{2}{*}{Empty node improved baseline, single mT5-large model} & \textcolor{blue!18.6!black}{+0.66} & \textcolor{blue!16.5!black}{+0.2} & \textcolor{blue!16.5!black}{+0.7} & \textcolor{red!0.0!black}{+0.0} & \textcolor{blue!16.8!black}{+0.7} & \textcolor{blue!33.8!black}{+1.9} & \textcolor{blue!23.0!black}{+0.2} & \textcolor{blue!28.5!black}{+1.4} & \textcolor{blue!28.2!black}{+0.9} & \textcolor{blue!13.1!black}{+0.1} & \textcolor{blue!8.1!black}{+0.1} & \textcolor{blue!9.7!black}{+1.1} \\[-1pt]
~~ & \textcolor{black}{72.25} & \textcolor{black}{81.5} & \textcolor{black}{75.1} & \textcolor{black}{76.9} & \textcolor{black}{72.7} & \textcolor{black}{60.2} & \textcolor{black}{82.2} & \textcolor{black}{71.0} & \textcolor{black}{66.7} & \textcolor{black}{68.8} & \textcolor{black}{76.5} & \textcolor{black}{63.3} \\[.5pt]
~~\multirow{2}{*}{Gold empty nodes, single mT5-large model} & \textcolor{blue!100.0!black}{@+3.55} & \textcolor{blue!100.0!black}{@+0.8} & \textcolor{blue!100.0!black}{@+4.5} & \textcolor{blue!100.0!black}{@+1.5} & \textcolor{blue!100.0!black}{@+4.3} & \textcolor{blue!100.0!black}{@+5.5} & \textcolor{blue!100.0!black}{@+0.7} & \textcolor{blue!100.0!black}{@+5.0} & \textcolor{blue!100.0!black}{@+3.1} & \textcolor{blue!100.0!black}{@+0.6} & \textcolor{blue!100.0!black}{@+1.7} & \textcolor{blue!100.0!black}{@+11.1} \\[-1pt]
~~ & \textcolor{black}{@75.14} & \textcolor{black}{@82.1} & \textcolor{black}{@78.9} & \textcolor{black}{@78.4} & \textcolor{black}{@76.3} & \textcolor{black}{@63.8} & \textcolor{black}{@82.7} & \textcolor{black}{@74.6} & \textcolor{black}{@68.9} & \textcolor{black}{@69.3} & \textcolor{black}{@78.1} & \textcolor{black}{@73.3} \\[.5pt]
    \midrule
\multicolumn{13}{l}{\textsc{B) Empty Node Prediction Ablations for umT5-xl Model}} \\[2pt]
~~Empty node baseline, single umT5-xl model & \textcolor{black}{74.65} & \textcolor{black}{83.8} & \textcolor{black}{77.4} & \textcolor{black}{80.8} & \textcolor{black}{74.3} & \textcolor{black}{62.0} & \textcolor{black}{84.0} & \textcolor{black}{73.7} & \textcolor{black}{67.5} & \textcolor{black}{73.4} & \textcolor{black}{79.3} & \textcolor{black}{64.8} \\[.5pt]
~~\multirow{2}{*}{No empty nodes, single umT5-xl model} & \textcolor{red!100.0!black}{--\kern 0.04em 12.6} & \textcolor{red!100.0!black}{--\kern 0.04em 6.4} & \textcolor{red!100.0!black}{--\kern 0.04em 8.7} & \textcolor{red!100.0!black}{--\kern 0.04em 4.7} & \textcolor{red!100.0!black}{--\kern 0.04em 16} & \textcolor{red!100.0!black}{--\kern 0.04em 13} & \textcolor{red!100.0!black}{--\kern 0.04em 6.2} & \textcolor{red!100.0!black}{--\kern 0.04em 17} & \textcolor{red!100.0!black}{--\kern 0.04em 15} & \textcolor{red!100.0!black}{--\kern 0.04em 8.2} & \textcolor{red!100.0!black}{--\kern 0.04em 12} & \textcolor{red!100.0!black}{--\kern 0.04em 31} \\[-1pt]
~~ & \textcolor{black}{62.10} & \textcolor{black}{77.4} & \textcolor{black}{68.7} & \textcolor{black}{76.1} & \textcolor{black}{57.7} & \textcolor{black}{49.4} & \textcolor{black}{77.8} & \textcolor{black}{57.0} & \textcolor{black}{53.4} & \textcolor{black}{65.2} & \textcolor{black}{67.0} & \textcolor{black}{33.6} \\[.5pt]
~~\multirow{2}{*}{One-stage umT5-xl model} & \textcolor{blue!19.1!black}{+0.74} & \textcolor{red!0.8!black}{+0.0} & \textcolor{blue!19.5!black}{+1.0} & \textcolor{red!12.3!black}{--\kern 0.04em 0.6} & \textcolor{red!0.5!black}{--\kern 0.04em 0.1} & \textcolor{blue!52.7!black}{+3.1} & \textcolor{red!0.0!black}{+0.0} & \textcolor{blue!40.3!black}{+1.5} & \textcolor{blue!41.0!black}{+2.0} & \textcolor{red!15.3!black}{--\kern 0.04em 1.2} & \textcolor{blue!13.7!black}{+0.3} & \textcolor{blue!18.2!black}{+2.3} \\[-1pt]
~~ & \textcolor{black}{75.39} & \textcolor{black}{83.8} & \textcolor{black}{78.3} & \textcolor{black}{80.2} & \textcolor{black}{74.2} & \textcolor{black}{65.1} & \textcolor{black}{84.0} & \textcolor{black}{75.2} & \textcolor{black}{69.5} & \textcolor{black}{72.2} & \textcolor{black}{79.6} & \textcolor{black}{67.1} \\[.5pt]
~~\multirow{2}{*}{Empty node improved baseline, single umT5-xl model} & \textcolor{blue!13.2!black}{+0.51} & \textcolor{blue!3.5!black}{+0.0} & \textcolor{blue!6.1!black}{+0.3} & \textcolor{red!0.0!black}{+0.0} & \textcolor{blue!12.2!black}{+0.5} & \textcolor{blue!14.3!black}{+0.9} & \textcolor{blue!6.5!black}{+0.1} & \textcolor{blue!46.2!black}{+1.7} & \textcolor{blue!28.8!black}{+1.4} & \textcolor{red!1.5!black}{--\kern 0.04em 0.1} & \textcolor{blue!4.0!black}{+0.1} & \textcolor{blue!6.3!black}{+0.8} \\[-1pt]
~~ & \textcolor{black}{75.16} & \textcolor{black}{83.8} & \textcolor{black}{77.7} & \textcolor{black}{80.8} & \textcolor{black}{74.8} & \textcolor{black}{62.9} & \textcolor{black}{84.1} & \textcolor{black}{75.4} & \textcolor{black}{68.9} & \textcolor{black}{73.3} & \textcolor{black}{79.4} & \textcolor{black}{65.6} \\[.5pt]
~~\multirow{2}{*}{Gold empty nodes, single umT5-xl model} & \textcolor{blue!100.0!black}{@+3.87} & \textcolor{blue!100.0!black}{@+0.9} & \textcolor{blue!100.0!black}{@+5.0} & \textcolor{blue!100.0!black}{@+1.9} & \textcolor{blue!100.0!black}{@+4.4} & \textcolor{blue!100.0!black}{@+5.9} & \textcolor{blue!100.0!black}{@+0.8} & \textcolor{blue!100.0!black}{@+3.7} & \textcolor{blue!100.0!black}{@+4.9} & \textcolor{blue!100.0!black}{@+0.8} & \textcolor{blue!100.0!black}{@+1.8} & \textcolor{blue!100.0!black}{@+12.7} \\[-1pt]
~~ & \textcolor{black}{@78.52} & \textcolor{black}{@84.7} & \textcolor{black}{@82.4} & \textcolor{black}{@82.7} & \textcolor{black}{@78.7} & \textcolor{black}{@67.9} & \textcolor{black}{@84.8} & \textcolor{black}{@77.4} & \textcolor{black}{@72.3} & \textcolor{black}{@74.2} & \textcolor{black}{@81.1} & \textcolor{black}{@77.5} \\[.5pt]
    \midrule
\multicolumn{13}{l}{\textsc{C) Empty Node Prediction Ablations for umT5-xxl Model}} \\[2pt]
~~Empty node baseline, single umT5-xxl model & \textcolor{black}{75.61} & \textcolor{black}{84.0} & \textcolor{black}{78.5} & \textcolor{black}{81.3} & \textcolor{black}{74.9} & \textcolor{black}{65.6} & \textcolor{black}{84.6} & \textcolor{black}{77.9} & \textcolor{black}{66.5} & \textcolor{black}{71.8} & \textcolor{black}{80.8} & \textcolor{black}{65.7} \\[.5pt]
~~\multirow{2}{*}{No empty nodes, single umT5-xxl model} & \textcolor{red!100.0!black}{--\kern 0.04em 12.6} & \textcolor{red!100.0!black}{--\kern 0.04em 6.7} & \textcolor{red!100.0!black}{--\kern 0.04em 8.9} & \textcolor{red!100.0!black}{--\kern 0.04em 5.0} & \textcolor{red!100.0!black}{--\kern 0.04em 17} & \textcolor{red!100.0!black}{--\kern 0.04em 13} & \textcolor{red!100.0!black}{--\kern 0.04em 6.6} & \textcolor{red!100.0!black}{--\kern 0.04em 18} & \textcolor{red!100.0!black}{--\kern 0.04em 15} & \textcolor{red!100.0!black}{--\kern 0.04em 7.2} & \textcolor{red!100.0!black}{--\kern 0.04em 12} & \textcolor{red!100.0!black}{--\kern 0.04em 31} \\[-1pt]
~~ & \textcolor{black}{63.05} & \textcolor{black}{77.3} & \textcolor{black}{69.6} & \textcolor{black}{76.4} & \textcolor{black}{58.2} & \textcolor{black}{53.3} & \textcolor{black}{78.0} & \textcolor{black}{60.1} & \textcolor{black}{52.2} & \textcolor{black}{64.6} & \textcolor{black}{69.0} & \textcolor{black}{34.5} \\[.5pt]
~~\multirow{2}{*}{One-stage umT5-xxl model} & \textcolor{blue!18.7!black}{+0.74} & \textcolor{blue!48.3!black}{+0.3} & \textcolor{blue!11.7!black}{+0.7} & \textcolor{blue!6.0!black}{+0.1} & \textcolor{blue!35.9!black}{+1.6} & \textcolor{blue!72.0!black}{+4.5} & \textcolor{blue!24.1!black}{+0.2} & \textcolor{blue!80.5!black}{+1.6} & \textcolor{red!3.7!black}{--\kern 0.04em 0.5} & \textcolor{red!19.9!black}{--\kern 0.04em 1.4} & \textcolor{red!3.0!black}{--\kern 0.04em 0.4} & \textcolor{blue!10.8!black}{+1.6} \\[-1pt]
~~ & \textcolor{black}{76.35} & \textcolor{black}{84.3} & \textcolor{black}{79.2} & \textcolor{black}{81.5} & \textcolor{black}{76.5} & \textcolor{black}{70.1} & \textcolor{black}{84.8} & \textcolor{black}{79.5} & \textcolor{black}{66.0} & \textcolor{black}{70.3} & \textcolor{black}{80.4} & \textcolor{black}{67.2} \\[.5pt]
~~\multirow{2}{*}{Empty node improved baseline, single umT5-xxl model} & \textcolor{blue!13.5!black}{+0.53} & \textcolor{blue!5.2!black}{+0.1} & \textcolor{blue!15.2!black}{+0.9} & \textcolor{red!0.0!black}{+0.0} & \textcolor{blue!14.1!black}{+0.6} & \textcolor{blue!10.3!black}{+0.7} & \textcolor{blue!8.4!black}{+0.0} & \textcolor{blue!47.0!black}{+0.9} & \textcolor{blue!26.7!black}{+1.3} & \textcolor{red!0.4!black}{+0.0} & \textcolor{red!0.5!black}{--\kern 0.04em 0.1} & \textcolor{blue!10.9!black}{+1.6} \\[-1pt]
~~ & \textcolor{black}{76.14} & \textcolor{black}{84.1} & \textcolor{black}{79.4} & \textcolor{black}{81.3} & \textcolor{black}{75.5} & \textcolor{black}{66.3} & \textcolor{black}{84.6} & \textcolor{black}{78.8} & \textcolor{black}{67.8} & \textcolor{black}{71.8} & \textcolor{black}{80.7} & \textcolor{black}{67.2} \\[.5pt]
~~\multirow{2}{*}{Gold empty nodes, single umT5-xxl model} & \textcolor{blue!100.0!black}{@+4.00} & \textcolor{blue!100.0!black}{@+0.6} & \textcolor{blue!100.0!black}{@+5.8} & \textcolor{blue!100.0!black}{@+2.0} & \textcolor{blue!100.0!black}{@+4.5} & \textcolor{blue!100.0!black}{@+6.2} & \textcolor{blue!100.0!black}{@+0.8} & \textcolor{blue!100.0!black}{@+2.0} & \textcolor{blue!100.0!black}{@+4.7} & \textcolor{blue!100.0!black}{@+1.3} & \textcolor{blue!100.0!black}{@+1.7} & \textcolor{blue!100.0!black}{@+14.7} \\[-1pt]
~~ & \textcolor{black}{@79.61} & \textcolor{black}{@84.6} & \textcolor{black}{@84.3} & \textcolor{black}{@83.4} & \textcolor{black}{@79.4} & \textcolor{black}{@71.8} & \textcolor{black}{@85.4} & \textcolor{black}{@79.9} & \textcolor{black}{@71.2} & \textcolor{black}{@73.1} & \textcolor{black}{@82.5} & \textcolor{black}{@80.3} \\[.5pt]
    \bottomrule
    \end{tabular*}

  \caption{Ablation experiments on the CorefUD 1.4 minidev set for datasets that contain empty nodes (CoNLL score in \%). The results of submitted models are averages of 7 runs, other results are averages of 2 or more runs.}
  \label{tab:dev_enodes_ablations}
\end{table*}

\paragraph{Effect of Empty Node Quality}\looseness-1
Finally, Table~\ref{tab:dev_enodes_ablations} shows the effect of empty node
prediction quality on the overall coreference resolution performance for every
dataset with empty nodes. Compared to the submitted models, not using any empty
node predictions results in a drop of 12.3 percent points on average, while
using gold empty nodes provides an average boost of 3.8 percent points. While
using the improved empty node prediction system improves the results by 0.55
percent points on average, the best results are achieved by the one-stage
variant delivering an average increase of 0.7 percent points compared to
the standard two-stage variant.

\section{Conclusions}

We introduced CorPipe 26, the winning submission to the \CRAClong~\citep{novak-etal-2026-findings}.
Our approach encompasses two variants. The first is a two-stage pipeline
architecture that first predicts empty nodes using a dedicated pretrained
encoder model, and then performs mention detection and coreference linking through
a jointly trained system utilizing another pretrained encoder. The second
variant predicts empty nodes together with mentions and coreference links in
a single model. Our system significantly outperforms all other submissions in
both the LLM track and the unconstrained track by 2.8 and 9.5 percent
points, respectively. The source code and trained models are publicly available at
\hbox{\small\url{https://github.com/ufal/crac2026-corpipe}}.

\paragraph{Future Work}

While CorPipe has been the top-performing system in the CRAC Shared Tasks for
the past years, there is always room for improvement. First, CorPipe does
not handle coreference links longer than the maximum segment length (by default
2\,560 tokens). Our initial attempts to address this
issue by linking mentions across segments using the current models were
unsuccessful, but modifying the training procedure to include cross-segment
links could help.

Second, CorPipe currently relies on pretrained \textit{encoder}
models. Given the abundance of new pretrained decoder-only models compared
to new multilingual encoders, exploring decoder-only architectures is
a possible future direction.

Last, making CorPipe available as a user-friendly tool or a web service
could facilitate its adoption by the research community and practitioners.

\section*{Acknowledgements}

Our research has been supported by the OP~JAK project
CZ.02.01.01/00/23\_020/0008518 of the Ministry of Education, Youth and Sports
of the Czech Republic and uses data provided by the \hbox{LINDAT/CLARIAH-CZ}
\hbox{Research} Infrastructure (https://lindat.cz), supported by the Ministry
of Education, Youth and Sports of the Czech Republic (Project No. LM2023062).

\bibliography{anthology,custom}

\appendix
\clearpage
\onecolumn

\section{Improved Empty Node Prediction Baseline}
\label{sec:appendix_empty_nodes}

\begin{table}[h]
  \footnotesize  \setlength{\tabcolsep}{3pt}
  \begin{tabular*}{\hsize}{l*{11}{@{\extracolsep{\fill}}c}}
    \toprule
      Treebank & ARC & DEP & WO & DEP\_WO & FORM & LEMMA & UPOS & XPOS & FEATS & ALL \\
    \midrule
      \texttt{ca}              & 96.73 & 96.73 & 93.93 & 93.93 & ---   & ---   & ---   & ---   & ---   & 93.93 \\
      \texttt{cs\_pcedt}       & 76.86 & 74.84 & 76.74 & 74.72 & 74.21 & 76.11 & 76.74 & 76.49 & 74.08 & 73.58 \\
      \texttt{cs\_pdt}         & 78.06 & 77.44 & 77.75 & 77.13 & 75.88 & 76.50 & 78.06 & 77.60 & 76.03 & 75.41 \\
      \texttt{cs\_pdtsc}       & 88.63 & 87.54 & 87.89 & 86.85 & 86.45 & 87.79 & 88.53 & 88.08 & 86.11 & 85.27 \\
      \texttt{cu}              & 83.37 & 82.97 & 82.58 & 82.32 & ---   & ---   & ---   & ---   & ---   & 82.32 \\
      \texttt{es}              & 96.09 & 96.09 & 93.82 & 93.82 & ---   & ---   & ---   & ---   & ---   & 93.82 \\
      \texttt{grc}             & 91.54 & 89.88 & 91.54 & 89.88 & ---   & ---   & ---   & ---   & ---   & 89.88 \\
      \texttt{hu\_korkor}      & 91.45 & 88.16 & 91.45 & 88.16 & 91.45 & ---   & ---   & ---   & ---   & 88.16 \\
      \texttt{hu\_szegedkoref} & 92.46 & 89.80 & 92.02 & 89.36 & ---   & ---   & ---   & ---   & ---   & 89.36 \\
      \texttt{pl}              & 90.67 & 90.55 & 90.67 & 90.55 & ---   & ---   & ---   & ---   & ---   & 90.44 \\
      \texttt{tr}              & 86.84 & 86.84 & 86.84 & 86.84 & 84.40 & ---   & ---   & ---   & ---   & 84.40 \\
    \bottomrule
  \end{tabular*}

  \caption{Performance of the improved empty nodes prediction system on the
  minidev sets of the CRAC 2026 dataset containing empty nodes. Each reported
  metric is an F1 score where a prediction is considered correct if both the
  dependency head and the given column are correct.}
  \label{tab:empty_nodes_improved}
\end{table}

\end{document}